\documentclass{article}

\usepackage{arxiv}

\usepackage[utf8]{inputenc} 
\usepackage[T1]{fontenc}    
\usepackage{hyperref}       
\usepackage{url}            
\usepackage{booktabs}       
\usepackage{amsmath}
\usepackage{amsfonts}       
\usepackage{nicefrac}       
\usepackage{microtype}      
\usepackage{lipsum}		
\usepackage{graphicx}
\usepackage{natbib}
\usepackage{doi}

\usepackage{framed}
\usepackage{multirow}

\usepackage{amssymb}
\usepackage{mathrsfs}
\usepackage{rotating}
\usepackage{appendix}
\usepackage{textcomp}
\usepackage{manyfoot}
\usepackage{etoolbox}
\usepackage{algorithm}
\usepackage{xcolor}
\usepackage{algorithmicx}
\usepackage{algpseudocode}
\usepackage{program}
\usepackage{listings}
\usepackage{breakurl}
\usepackage{geometry}
\usepackage{hypcap}
\usepackage{breakurl}
\usepackage{natbib}
\usepackage{wrapfig}
\usepackage{booktabs}
\usepackage{tabularx}
\usepackage[utf8]{inputenc}
\usepackage{hyperref}
\DeclareUnicodeCharacter{2212}{-}

\title{Video-based assessment of intraoperative surgical skill}

\author{ Sanchit Hira\thanks{These authors contributed equally to this work.} \\
	Laboratory for Computational Sensing \& Robotics\\
	Johns Hopkins University\\
	Baltimore, MD, USA \\
	\AND
	Digvijay Singh\thanks{These authors contributed equally to this work.} \\
	Department of Computer Science\\
	Johns Hopkins University\\
	Baltimore, MD, USA \\
    \AND
	Tae Soo Kim\thanks{These authors contributed equally to this work.} \\
	Department of Computer Science\\
	Johns Hopkins University\\
	Baltimore, MD, USA \\
    \AND
	Shobhit Gupta \\
	Department of Computer Science\\
	Johns Hopkins University\\
	Baltimore, MD, USA \\
	\AND
	Gregory Hager \\
	Department of Computer Science\\
	Malone Center for Engineering in Healthcare \\
	Johns Hopkins University\\
	Baltimore, MD, USA \\
    \AND
	Shameema Sikder \\
	Wilmer Eye Institute \\
	Johns Hopkins University School of Medicine\\
	Baltimore, MD, USA \\
    \AND
	S.~Swaroop Vedula \\
	Malone Center for Engineering in Healthcare\\
	Johns Hopkins University\\
	Baltimore, MD, USA \\
	\texttt{swaroop@jhu.edu} \\
}

\renewcommand{\shorttitle}{Video Based Assessment of surgical skill}

\begin{document}
\maketitle

\begin{abstract}
	\textbf{Purpose:} The objective of this investigation is to provide a comprehensive analysis of state-of-the-art methods for video-based assessment of surgical skill in the operating room.

    \textbf{Methods:} Using a data set of 99 videos of capsulorhexis, a critical step in cataract surgery, we evaluate feature based methods previously developed for surgical skill assessment mostly under benchtop settings. In addition, we present and validate two deep learning methods that directly assess skill using RGB videos. In the first method, we predict instrument tips as keypoints, and learn surgical skill using temporal convolutional neural networks. In the second method, we propose a novel architecture for surgical skill assessment that includes a frame-wise encoder (2D convolutional neural network) followed by a temporal model (recurrent neural network), both of which are augmented by visual attention mechanisms. We report the area under the receiver operating characteristic curve, sensitivity, specificity, and predictive values with each method through 5-fold cross-validation.

    \textbf{Results:} For the task of binary skill classification (expert vs. novice), deep neural network based methods exhibit higher AUC than the classical spatiotemporal interest point based methods. The neural network approach using attention mechanisms also showed high sensitivity and specificity.

    \textbf{Conclusion:} Deep learning methods are necessary for video-based assessment of surgical skill in the operating room. Our findings of internal validity of a network using attention mechanisms to assess skill directly using RGB videos should be evaluated for external validity in other data sets.
\end{abstract}

\keywords{video-based assessment \and surgical skill \and deep learning \and cataract surgery}

\section{Introduction}
Surgeons' skill in the operating room affects patient outcomes \citep{birkmeyer2013surgical, nathan2012surgical, nathan2014technical, curtis2020association}. This means that interventions to optimize surgeons' skill can potentially improve quality of patient care. Assessment of skill is a cornerstone for any intervention to improve it. In addition, surgical skill assessment is essential for several purposes throughout surgeons' career including in-training evaluations, surgical coaching, certification, re-certification, and credentialing of surgeons, as well as end of career decisions \citep*{maier2017surgical}.

Traditionally, surgical skill assessment was based upon direct observation of surgeons in the operating room \citep{vedula2017objective}. But this method provides a subjective opinion of the assessing surgeon. As a result, the assessment is unreliable despite use of structured rating scales such as the objective structured assessment of surgical technical skill (OSATS) global rating scale. Unlike direct observation, video-based assessment enables asynchronous evaluation of surgeons' skill. In addition, video-based assessment can be used for providing surgeons with formative assessments, coaching surgeons, among other applications \citep{valanci2020implementation, pugh2020}.

Currently, video-based assessment of surgical skill in the operating room is obtained either from ratings by peer surgeons or crowd raters (i.e., crowdsourcing) \citep{pugh2020}. Self-assessment by surgeons is not accurate, particularly, among those with poor skill \citep{pandey2008self}. Crowd-sourced skill ratings correlate with expert ratings \citep{lendvay2015crowdsourcing,malpani2015study,powers2016crowdsourcing}, but some studies show low predictive value for the ratings \citep{aghdasi2015crowd,deal2017evaluation}. Consequently, despite the efficiency with which skill ratings can be obtained with crowdsourcing, its role in routine assessments of surgical skill is not established. However, video-based assessment of surgical skill, either by peer surgeons or by crowd raters, is inherently subjective. Furthermore, video-based assessment by peers relies upon access to experienced raters. Thus, manual video-based assessment is a subjective and expensive approach to provide surgeons with skill assessments on a routine basis.

Surgical data science methods can address limitations of subjectivity and inefficiency in manual skill assessment through quantitative analysis of videos of the surgical field \citep{maier2017surgical}. The conventional approach to assess surgical skill using data was based upon instrument motion \cite{ahmidi2017dataset}. Although some methods to analyze surgical videos were developed, they were limited to the simulation setting and not adequately validated \citep{zia2018automatedpaper,bettadapura2013augmenting,zia2015automated,sharma2014surgical,zia2018video,tao2012sparse,ahmidi2017dataset}. In fact, the majority of previous studies on surgical skill assessment using instrument motion or video data were limited to the simulation setting. Thus, despite their promise to address surgeons' need for efficient, reliable, and unbiased assessments of their skill, surgical data science methods to assess surgical skill in the operating room are not sufficiently developed.

Previously, we developed and evaluated a temporal convolutional neural network to assess skill using manually annotated instrument tips in video images \citep{kim2019objective}. Our objective in this study is to develop and validate methods for assessment of surgical skill directly from videos of the surgical field. Using a dataset of 99 videos, we evaluated all existing algorithms to assess surgical skill with a uniform cross-validation setup and evaluation metrics. For skill assessment in the operating room using videos of the surgical field, to our knowledge, this dataset is the largest in the literature. The contributions of our work described in this paper are as follows:
\begin{enumerate}
    \item An exhaustive evaluation of feature based methods to assess surgical skill in the operating room;
    \item Evaluation of methods to directly analyze RGB videos to provide a surgical skill assessment;
    \item A novel architecture for a deep learning method augmented with attention to assess surgical skill, towards explaining the predicted skill label / score.
\end{enumerate}

\section{Methods}\label{sec2}
In this section, we describe a comprehensive list of existing feature based approaches for surgical skill assessment considered in this work. Then, we present our tool detection based approach and a dual attention based approach for skill assessment.

\subsection{Feature based approaches}
\label{sec:cvmethods}
We evaluate five approaches to obtain features from videos, which we then analyze with a linear support vector machine. The five approaches for feature extraction are described below.

\subsubsection{Detectors (STIPs)}
\label{sec:stips}
Following \citep{laptev2005space}, we use STIPs to identify regions in images that are key to developing useful feature representations.
The video is convolved with a 3D Gaussian kernel to construct a $3x3x3$ spatiotemporal second order moment matrix, which is composed of first order spatial and temporal derivatives averaged using a Gaussian weighing function. Gaussian smoothing is performed on this response function followed by non-maximal suppression.

\subsubsection{Descriptors}
\label{sec:descriptors}
We compute three different descriptors for each interest point detected using the approach described in Section \ref{sec:stips}: histogram of oriented gradients (HoG) to encode local spatial information in image patches \citep{dalal2006human}, histogram of optical flow (HoF) to describe localized flow in videos \citep{dalal2006human}, and motion boundary histograms (MBHs) to encode localized temporal information in X and Y components of the differential flow \citep{zia2016automated}.

\subsubsection{Features}

\paragraph{\textbf{Bag of Words (BoW)}}
\label{sec:bow}
For BoW, we cluster descriptors of interest points in a video using a k-means algorithm to obtain a visual feature vocabulary, and use TF-IDF to compute the BoW feature \citep{robertson2004understanding}.

\paragraph{\textbf{Augmented Bag of Words (Aug. BoW)}}
The BoW feature lacks temporal information, which we find is important in assessing surgical skill. We first create a temporal vocabulary by classifying the time between interest points into N bins. We concatenate the visual and temporal vocabularies and compute n-grams. The n-grams are thus a sequence of events and sum of their times. This is a useful representation for surgical movements that take relatively longer periods of time and continue over a number of frames. We use TF-IDF to represent n-grams \citep{bettadapura2013augmenting}.

\paragraph{\textbf{Discrete Fourier Transform (DFT) / Discrete Cosine Transform (DCT)}}
DFT and DCT extract motion information i.e. the frequencies of the different surgical action cluster categories [cite]. A transformation matrix or motion class matrix (MCM) is created from the learned clusters of the visual vocabulary using K-means, so that if a video has N frames, a KxN matrix is created that contains for each row i = 1,2, …, K, the counts of how many interest points in the nth frame belong to the cluster K. The DFT and DCT of this matrix are then calculated, where each entry in the calculated KxN matrix represents the nth frequency coefficient of the kth cluster. This matrix measures the repetitiveness of the different surgical actions in the videos. Since higher frequencies are a result of noisy or abrupt movements, the lowest D frequencies are used, and we get a KxD matrix.  This matrix is flattened to get a K*D feature vector.

\paragraph{\textbf{Sequential Motion Textures (SMT)}}
\label{sec:smt}
In SMT, spatiotemporal information is encoded in the form of gray level co-occurrence matrices (GLCMs) that are derived from frame kernel matrices of the MCM \citep{sharma2014surgical}. They represent the affinity of the elements of a matrix with respect to all the elements. The frame kernel matrices are obtained by splitting the MCM into time windows of specified width (W), which allows temporal information to be considered over a fixed time interval according to the length of the videos. This is followed by applying a radial basis frequency kernel and shifting the domain to grayscale. Texture patterns are then extracted from these matrices using GLCMs for different gray levels Ng, which encode both spatial relationships and motion dynamics in the surgical videos. After calculating the averaged and normalized GLCMs for the windows, 20 standard texture features are selected using sequential forward feature selection; this results in Wx20 features for each video \citep{marcano2010feature}.

\paragraph{\textbf{Approximate Entropy (ApEn) / Cross Approximate Entropy (XApEn)}}
\label{sec:apen}
ApEn and XApEn are methods that construct entropy-based features that measure the amount of entropy in a given time-series input \citep{zia2018video}. Authors of \citep{zia2018video} claim that entropy-based features help recognize predictable and regular patterns in time-series which in turn leads to better skill assessment from video-based descriptors and accelerometer data.

For ApEn, for each time series belonging to a motion class, it is dependent on the emdedding dimension m, radius r and time delay tau. Each time series is split into embedding factors (time windows) according to time delay, and then for each embedding factor the frequency of repeatable patterns is calculated from by summing up the Heaviside functions of the L-Infinity norm distance MCMs (PAPER). XApEn calculation is similar to ApEn, except that the the L-Infinity norm distance is calculated between embedding factors for time series for cluster pairs and is done for all the possible cluster pairs.

\subsection{Tool detection approach for surgical skill from videos}
\label{sec:toolmethods}
There exists evidence that a machine can learn to recognize surgical skill from tool motion data \citep{kim2019objective} by learning a temporal convolutional neural network (TCN) on crowd sourced tool tip locations. Using a crowd sourcing approach, tool annotations can be obtained at scale to curate a large training corpus. However, an approach proposed in \citep{kim2019objective} requires the same level of annotation at test time which means that assessment is impossible for video samples without tool annotations. We address this limitation by learning to detect tool tips as keypoints of objects from video and analyze the predicted keypoints using a TCN (KP-TCN). This method thus allows us to accurately assess skill from video without requiring tool level annotation during inference.

\subsubsection{Learning to detect surgical instrument tips}
\label{sec:tools}
Given a trajectory of tool tip locations, Kim, et.al. \citep{kim2019objective} use a temporal convolutional neural network to model skill. In our pipeline, we include an additional stage to infer tool tip locations directly from images.

We model surgical tool tip locations as keypoints of objects. 
Let $X=\{x_1,x_2,\dots,x_N\}$ be a video with $N$ images. We assume there exists annotations regarding tool locations $y_n=\{p_1^n,\dots,p_K^n\}$ for each frame $x_n$ where $p_k^n \in \mathbb{R}^{2}$ is the pixel location of the $k$-th keypoint in frame $n$. Then, we wish to learn a keypoint detector $\mathcal{F}$ that minimizes the following objective for all available training videos with tool annotations:
\begin{equation}
    \begin{split}
    \{q_1^n,\dots,q_K^n\} =& \mathcal{F}(x_n)  \\
    \mathcal{L}_{keypoints} =& \sum_{n=1}^N \sum_{k=1}^K d(q_k^n,p_k^n)
    \end{split}
\end{equation}
where $d$ is a distance between the keypoints and $q_k^n \in \mathbb{R}^2$ is the keypoint prediction given $x_n$. In this work, we use a convolutional neural network based keypoint detector for $\mathcal{F}$ and train it in an end-to-end manner.

\subsubsection{Temporal Convolutional Neural Networks}
Given a trained tool tip localization network $\mathcal{F}$, we use a temporal convolutional neural network (TCN) for skill assessment presented in \citep{kim2019objective} to predict skill from videos. Using $\mathcal{F}$, we encode a video $X=\{x_1,x_2,\dots,x_N\}$ as a time-series of keypoint predictions $Z$ such that:
\begin{equation}
    Z = \{z_1,z_2,\dots,z_N\} \in \mathbb{R}^{N \times 2K}
\end{equation}
where $z_n = \mathcal{F}(x_n) \in \mathbb{R}^{2K}$ is the predicted pixel locations of $K$ keypoints in frame $x_n$. 

Compared to trajectories of tool locations, \citep{kim2019objective} has demonstrated that first derivative information of tools (tool velocities) is a better representation for skill assessment. Hence, we calculate tool velocities by considering the difference in predicted keypoint locations between successive frames.

The final input to the TCN is represented as $Z$ such that 
\begin{equation}
Z = \{ \delta z_1,\delta z_2 ,\dots,\delta z_N \}
\end{equation}
where $\delta z_n = z_{n+1}-z_n$ is the tool velocity at frame $n$. 

We follow the TCN design proposed in \cite{kim2019objective} which stacks layers of a temporal convolution layer followed by batch normalization operator (CITE BN) and an activation functions (ReLU). For the $l^{th}$ layer of the TCN with $F_l$ 1-D convolutional filters, the output activation can be written as
\begin{equation}
h^{l} = \sigma (W_l * h ^{l-1}) 
\end{equation}

where $h^{l}$ is the output of the $l$-th layer, $W_l$ are the convolutional filter weights of the $l^{th}$ layer, $\sigma(\cdot)$ is a non-linearity (ReLU followed by batch-normalization operation) and $*$ represents convolution. After the convolution operations, a global average pooling layer changes the size of the features in the temporal dimension to a fixed size, which is then fed into a linear layer for final classification. We train the TCN in an end-to-end manner using a standard cross entropy formulation.

\subsection{Dual attention network for skill assessment from videos (ATT)}
\label{sec:attention}
The TCN based approach assumes that the signal relevant to skill is embedded in the movements of the surgical instrument tip and learns to detect such patterns in an end-to-end manner using a convolutional neural network. In doing so, important contextual information such as visual changes in anatomy and instrument/anatomy interactions are omitted. In this section, we present an attention based method that performs skill assessment directly from RGB videos to address these issues. By using attention, the model implicitly attends to spatio-temporal regions of the video that is most relevant for the skill assessment task. 

Our architecture includes a feature extractor and a LSTM module, both equipped with separate but dependent spatial and temporal attention mechanisms. Various past studies have shown the ability of attention to generate better context vectors as well as to localize the relevant parts of the input for the task \citep{bahdanau2014neural}. We hypothesize that including attention in our model would enable us to localize time frames and locations in the video which might be useful for classification. Our detailed model architecture is presented below.

\subsubsection{Video Representation}
First, a ResNet encoder \citep{he2016deep} extracts the features from a given sequence of RGB frames, and then a LSTM cell \citep{hochreiter1997long} operates on those features to generate the final classification. We augment both the Resnet encoder and the LSTM Cell with attention \citep{xu2015show}, \citep{zhou2016attention}.

For the spatial domain, we remove the head from the ResNet base, such that the model outputs $D$ feature maps of size $H \times W$. All the image frames are passed through this base network, and a feature vector is obtained. The output of the feature extractor is represented by 
\begin{equation}
V = \{F_1, F_2, \dots F_N\}  , V \in \mathbb{R}^{N \times D \times H \times W}
\end{equation}
where $F_n \in \mathbb{R}^{D \times H \times W}$  is a spatial feature extracted from the spatial feature extractor and $N$ is the number of frames in video. 

Each of these frame features in turn consist of a $D$-dimensional positional activation map, which can be represented as
\begin{equation}
F_n = \{ a_{n1}, a_{n2}, \dots a_{nL} \}, \hspace{8pt} a_{ni} \in \mathbb{R}^D
\end{equation}
where $D$ is the dimension of the feature vector for each position in the image, and $L = H \times W$ is the number of positions in the image.

\subsubsection{Spatial and Temporal Attention}
After extraction, the features of each frame are sequentially passed through the LSTM cell. The hidden state of the LSTM cell from the previous frame is used to attend to the features of the current frame, using a 3 layer attention module.

The frame feature vectors $F_n$ are passed sequentially through the LSTM cell, and then both the hidden state of the LSTM cell and the frame feature vectors are used to compute the attention weights as follows

\begin{equation}
\begin{split}
e_{ni} = &f_{att}(F_n, \mathbf{h}_{n-1}) = f_{att}(\{ a_{n1}, a_{n2}, \dots a_{nL} \}, \mathbf{h}_{n-1}) \\
\alpha_{ni} = &\frac{\exp (e_{ni})}{\Sigma_{j=1}^L \exp (e_{nj})}  
\end{split}
\end{equation}

where $\alpha_{ni}$ are the positional attention weights, and $f_{att}$ is the spatial attention module, as described below. 
\begin{equation}
\begin{split}
att_f = &F_n \times W_{f} \\
att_l = &\mathbf{h}_{n-1} \times W_h \\
f_{att}(F_n, \mathbf{h}_{n-1}) = &ReLU(att_f + att_l) \times W_c
\end{split}
\end{equation}

where $W_f$, $W_h$ and $W_c$ represent linear layers. Once the attention weights $\alpha_{ni}$ are computed, the final context vector $z_n$ can be computed as

\begin{equation}
z_n = \Sigma_{i=1}^L \alpha_{ni} \cdot a_{ni}
\end{equation}

The attention weighted frame encoding vectors so generated are then fed into a temporal attention mechanism. First, a soft alignment score is computed between the output features of all the frames and the last hidden state of the LSTM cell using matrix multiplication. Softmax operation is applied to these alignment scores, which are then used as attention weights to compute the attended feature vector from the LSTM outputs.

\begin{equation}
    \begin{split}
         Z = &\{ z_1, z_2, \dots z_N \} \\
         M = &\mathbf{h}_{N} \times \tanh{(Z)} \\
        \beta = & softmax(M) \\         
        r = &H \beta
    \end{split}
\end{equation}

where $z_i$ is the spatial attention weighted frame encoding for frame $i$, $M$ are the soft alignment scores, $\beta$ are the temporal attention weights, $H$ is the combined LSTM output for all frames, and $r$ is the final temporal attention weighted encoding for the entire video.

Finally, a linear classification layer takes the attended feature encoding and generates the final classification. We use cross entropy loss and train the models for 1000 epochs with a step decay on plateau using stochastic gradient descent with an initial learning rate of $1e-2$. 

\section{Dataset} 
We used a dataset of 99 videos of capsulorhexis, which is a critical step in cataract surgery. The dataset is identical to that used in \citep{kim2019objective}. The Johns Hopkins Medical Institutions Institutional Review Board approved this retrospective study. Each instance of capsulorhexis in  the  dataset  was  completely  performed  by  one  surgeon.  A  faculty  surgeon  operated  in  28 instances and a trainee surgeon operated in 71 instances.  Videos of the surgical field were captured directly from the operating microscope; therefore, the videos did not include information identifying the surgeon or the patient. We processed each video to a resolution of 640*480 at 59 frames per second.

One expert surgeon watched each video and assigned a rating for skill using the International Council of Ophthalmology's Ophthalmology Surgical Competency Assessment Rubric for phacoemulsification (ICO-OSCAR:phaco) \citep{golnik2013development}. ICO-OSCAR:phaco includes two items to assess skill for capsulorhexis corresponding to surgical two goals in this step --- commencement of flap \& follow-through (CF), and capsulorhexis formation and completion (RF). Each item is rated on a scale of 2--5.

We specified the ground truth in two ways. First, we specified a binary skill class label --- expert / novice. Videos assigned a score of 5 on at least one item for capsulorhexis in ICO-OSCAR:phaco and at least a score of 4 on the other item were labeled expert; otherwise, the videos were labeled novice. Second, we specified a 3-class label for each item in ICO-OSCAR:phaco as a score of either 2 or 3, 4, or 5. 

For cross-validation purposes, we randomly split the data into five folds to minimize total duration of videos and to ensure a similar distribution of expert and novice skill labels across the folds. Table \ref{tab:datafolds} shows the distribution of videos in our dataset across the skill class labels.

\begin{table}
\centering
\caption{Distribution of groundtruth across cross-validation folds. \\
CF = Commencement of flap \& follow-through; \\
RF = Rhexis formation and completion}
\label{tab:datafolds}
\begin{tabular}{llllll}
\hline\noalign{\smallskip}
\textbf{Fold}   & \textbf{1}  & \textbf{2}  & \textbf{3}  & \textbf{4}  & \textbf{5}  \\
\hline\noalign{\smallskip}
\multicolumn{6}{l}{\textbf{Binary Skill Class}} \tabularnewline
\hline\noalign{\smallskip}
Expert & 10 & 10 & 10 & 10 & 12 \\
Novice & 9  & 9  & 9  & 9  & 11 \\
\noalign{\smallskip}\hline
\textbf{CF} & & & & & \\
\hline\noalign{\smallskip}
CF-2/3 & 4  & 1  & 3  & 4  & 2  \\
CF-4   & 8  & 9  & 6  & 6  & 10 \\
CF-5   & 7  & 9  & 10 & 9  & 11 \\
\hline\noalign{\smallskip}
\textbf{RF} & & & & & \\
\hline\noalign{\smallskip}
RF-2/3 & 6  & 4  & 5  & 5  & 3  \\
RF-4   & 3  & 6  & 6  & 6  & 12 \\
RF-5   & 10 & 9  & 8  & 8  & 8  \\
\noalign{\smallskip}\hline
\end{tabular}
\end{table}

\section{Experiments}
We performed two experiments -- predicting a binary skill class (expert / novice), and predicting a score on each of the two items for capsulorhexis in ICO-OSCAR:phaco (i.e., 3-class label for CF and RF). We used the same five fold cross-validation folds in the data for both the experiments (Table \ref{tab:datafolds}). Holding each fold out at a time for testing, we iteratively used one of the remaining four folds for validation and trained on the rest of the three folds. We used the model configuration and hyperparameters for which accuracy on the validation fold was highest.

\subsection{Implementation Detail and Settings}
\subsubsection{Feature based approaches}
To compute STIPs, we use windows of 6 seconds with a 2 second overlap on each side. The frame rate for input videos is 59 fps, resulting in 520 frames for each window. We use a 3x3x3 Gaussian kernel with spatial variance of 4 and temporal variance of 8 for the convolution. We use a Gaussian kernel with variance 1 to smooth the response function. The top 1000 STIPs are used for our experiments.

For HoGs, we extracted a 9x9x9 patch centered on each STIP, split it into 9 windows, and computed 8 gradient orientations resulting in a 72-dimensional feature vector. For HoFs, we used a 17x17x17 image patch centered on each STIP. We computed optical flow within the patch, and split it into 9 windows to compute 9 gradient orientations to obtain a 81-dimensional vector. For MBHs, we used 17x17 patches separately in the X and Y components of the optical flow that we split into 25 windows and compute 8 gradient orientations (HoGs). This gives a 400-dimensional vector (after concatenating the vector for X and Y components). We concatenated all three descriptors, resulting in a 513-dimensional vector, to compute features for each video. 

For k-means in Aug. BoW, we evaluated K = 25, 50, and 100 and using Euclidean and Mahalanobis distance. We computed n-grams using interspersed, cumulative, or pyramid encoding, where n = 3 and 5. In interspersed encoding, the temporal information is the time between events (or cluster labels), and this is encoded in sequence of the cluster labels occurring in the videos. This is useful for short surgical movements that take small amounts of time and are independent in the frames of the videos. Cumulative encoding encodes temporal information over a (user specified) sequence of events, and sums this up. Pyramid encoding creates l-grams for all of the specified n-grams from 1 to n. This can be interpreted as breaking down surgical movements and representing them at different levels. We used 5 bins to create the temporal vocabulary, following previous recommendation \citep{bettadapura2013augmenting,zia2018automatedpaper}.

For DFT and DCT, the frequencies for each cluster time series in the MCM are calculated, and the 50 lowest frequencies are selected from each to ignore noisy frequencies. This gives a $K*50$ vector for each of DFT and DCT, resulting in a 2xKx50 dimensional vector. The top 30 features are then selected using the SFFS feature selection method, yielding a final 30-dimensional frequency feature vector. SFFS works by evaluating the effectiveness of each individual feature using a classifier, and selecting the top n (specified by user) features.

For SMT, we analyzed multiple frame kernel matrix variances including 1e-8, 1e-7, 1e-6, 1e-5, 1e-4, and 1e-3. We evaluated multiple window sizes (W) of 2, 4, 8, and 16. We also analyzed multiple gray levels ($Ng$) of 8, 16, 32, 64, 128, and 256.

For entropy-based methods, we combined features from ApEn and XApEn. The dimensionality of features for ApEn is $r*K$, and for XApEn is $r*K*(K−1)/2$, where $K$ is the number of clusters, and ApEn and XApEn features are concatenated to produce a single feature vector. We have $\tau$ fixed at 1 (according to \citep{zia2018automatedpaper}), and $m$ taking values of 1 and 2. $r_{coeff}$ takes values 0.1, 0.15, 0.2, 0.25, where $r = r_{coeff} * \sigma$ where $\sigma$ is the standard deviation of the time series.

\subsection{Neural network approaches}
All neural network based approaches are implemented using the PyTorch \citep{pytorch} framework and trained using K80 GPUs. 

For tool detection, we train the HR-Net \citep{sun2019deep} using tool tip annotations of \citep{kim2019objective}. We use standard image augmentations during training which includes horizontal flip, shift-scale-rotate, brightness and Contrast adjustments. We follow the optimization proposed in \citep{sun2019deep} and use weighted focal loss \citep{focal} with a positive weight of 3 and 1-cycle learning rate schedule with an initial learning rate of 1e-2. We used the Adam optimizer for training.

For TCNs, we use an Adam \citep{adam} optimizer with a learning rate of 1e-3 and models train for 50 epochs. Given a time-series of tool tip predictions, we learn to predict skill class in an end-to-end fashion using standard cross entropy loss.

To train the dual attention network, we sample video snippets of 64 frames, sampled with a stride of 4 frames from each video. The image frames are normalized, and flip and rotate data augmentations are applied. The models are trained for 100 epochs with Adam \citep{adam} optimizer with an initial learning rate of 1e-2. We decrease the learning rate by a factor of 10 as validation loss plateaus. We again adopt a standard cross entropy loss.

At inference time using dual attention networks, a video is represented as clips of 64 frames sampled 4 frames apart to be consistent with training. The clips overlap by 32 frames. Predictions are made individually on all the clips and averaged to produce the final output.

We also evaluated a network trained with without attention (No ATT), using the same implementation described above for the dual attention network.

\subsection{Statistical Analyses}
To evaluate algorithm performance in each experiment, we computed the area under the receiver operating characteristic curve (AUC) as the primary metric, in addition to sensitivity, specificity, positive and negative predictive values (PPV, NPV), and accuracy (both micro- and macro-accuracy \cite{kim2019objective}). We used bootstrap to compute 95\% confidence intervals (CI) for AUC and computed the Wilson interval for the remaining measures \citep{agresti2003categorical}. We used the method described in \citep{hand2001simple} to estimate AUC and bootstrap to compute 95\% CI for alogrithms predicting a score on each item for capsulorhexis in ICO-OSCAR:phaco.

\section{Results}\label{sec3}
\begin{figure}
  \begin{center}
    \includegraphics[width=0.9\textwidth]{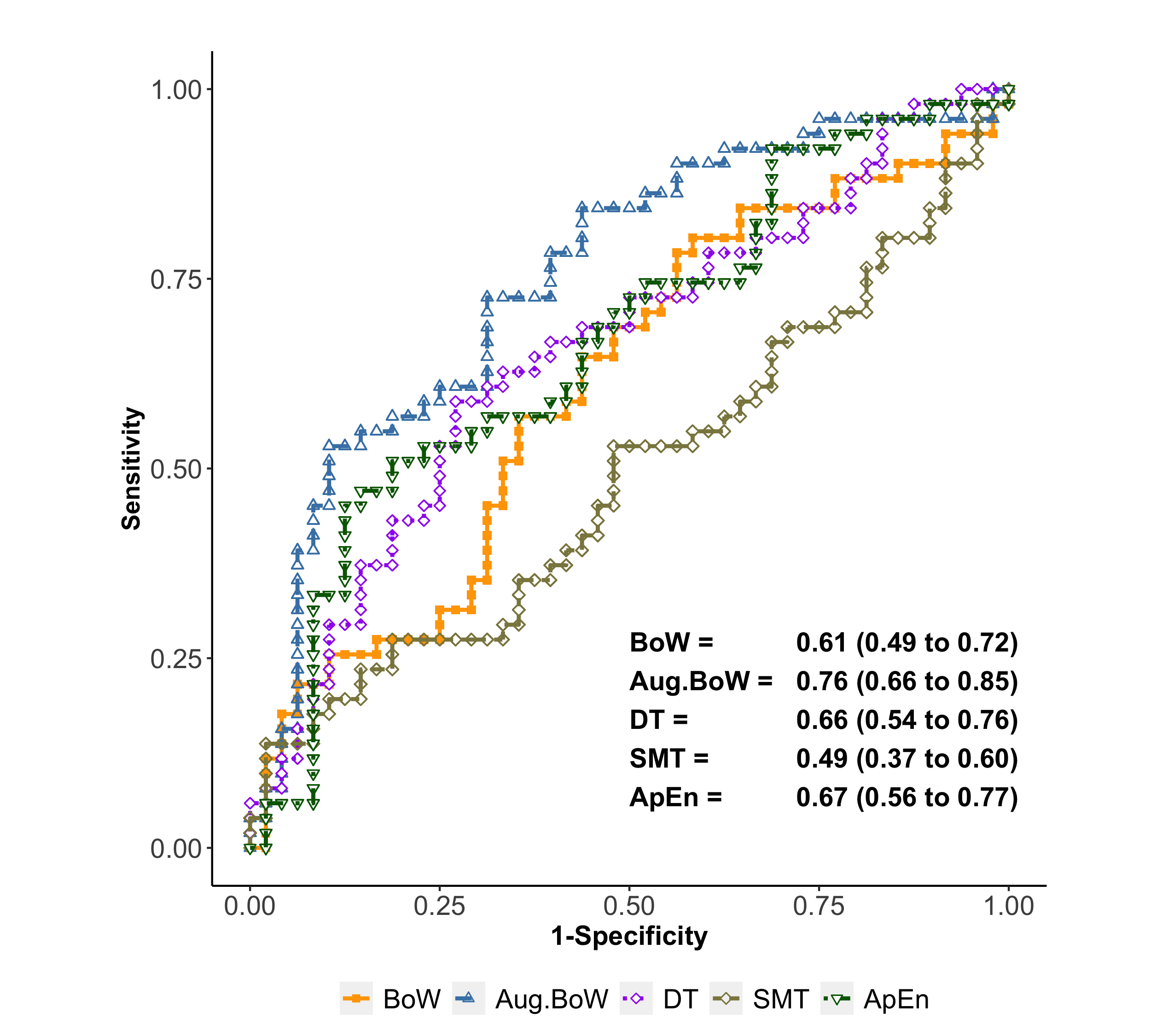}
  \end{center}
  \caption{ROC plots for interest-point based methods.\\
  Numbers on plots are AUC and 95\% confidence intervals.}
  \label{fig:roc_cvlinsvm}
\end{figure}

\paragraph{Feature based approaches} For predicting an expert / novice label using feature based methods, Figure \ref{fig:roc_cvlinsvm} shows receiver operating characteristic (ROC) curves and estimates of AUC. Of the 5 methods, the estimated AUC was highest for Aug. BoW. The 95\% CIs show that the estimated AUC was consistent with the null value (0.5) for BoW and SMT. For BoW and Aug. BoW the specificity was higher than sensitivity, whereas for DT, SMT, and ApEn sensitivity was higher than specificity (Figure \ref{fig:err_cvlinsvm} and Table 1 in Supplementary Material).

For predicting a 3-class label for CF and RF, estimates of AUC for all the feature based methods were consistent with the null value, although estimates of micro- and macro-accuracy, sensitivity, and specificity indicate that algorithm performance may have been affected by the class imbalance in our dataset (Tables 2 and 3 in Supplementary Material).

\begin{figure}
  \begin{center}
    \includegraphics[width=0.9\textwidth]{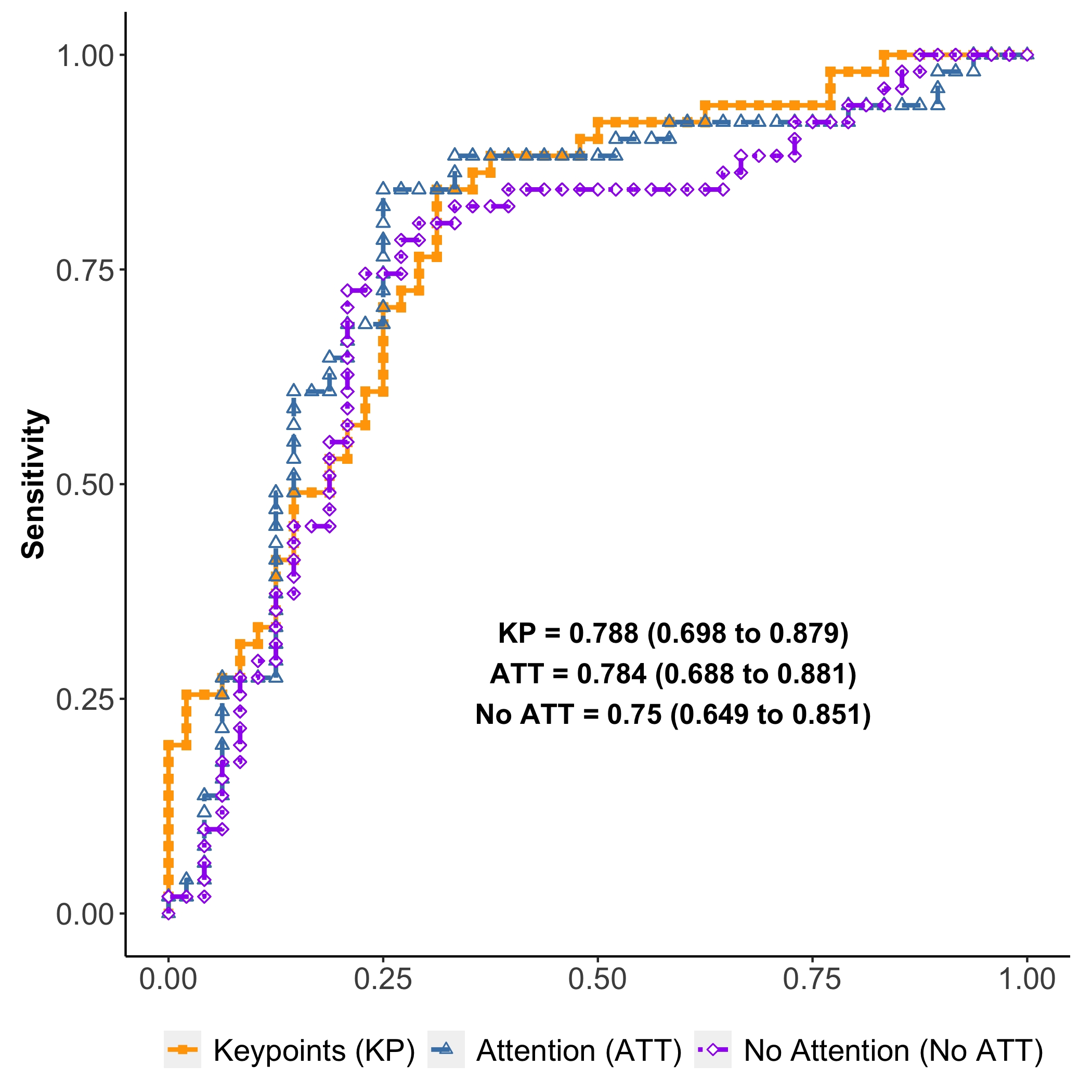}
  \end{center}
  \caption{ROC plots for deep learning methods.\\
  Numbers on plots are AUC and 95\% confidence intervals.}
  \label{fig:roc_dl}
\end{figure}

\begin{figure}
  \begin{center}
    \includegraphics[width=0.9\textwidth]{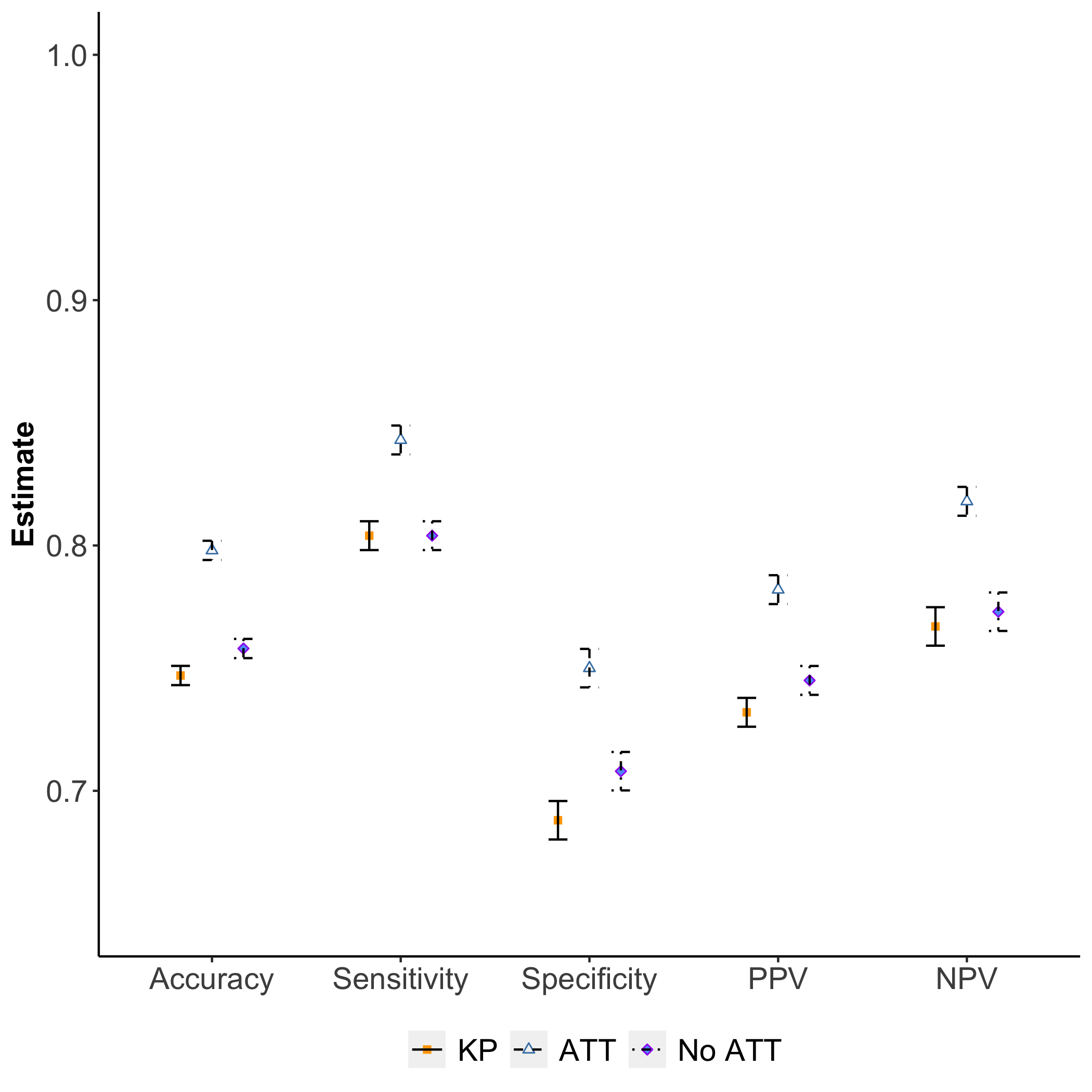}
  \end{center}
  \caption{Predictive performance of deep learning methods.}
  \label{fig:err_dl}
\end{figure}

\paragraph{Deep learning methods} For predicting an expert / novice label using deep learning methods, Figure \ref{fig:roc_dl} shows ROC curves and estimates of AUC. Attention mechanisms significantly improved algorithm performance. The attention-based network had high AUC, in addition to higher sensitivity and specificity than the other two deep learning methods (Figure \ref{fig:err_dl}). Although KP had similar AUC as the attention-based network, the sensitivity and specificity were lower for KP than the attention-based network.

For predicting a 3-class label for CF and RF using deep learning methods, estimates of AUC were lower than that for expert / novice label prediction (Table \ref{tab:dlacc}). In addition, these methods had higher sensitivity and lower specificity for labels indicating better skill, and lower sensitivity and higher specificity for labels indicating poor skill (Table \ref{tab:dlitems}). This observation, together with estimates of micro- and macro-accuracy suggest that performance of deep learning methods was susceptible to class imbalance in CF and RF in our dataset. 

\begin{figure}
  \begin{center}
    \includegraphics[width=0.9\textwidth]{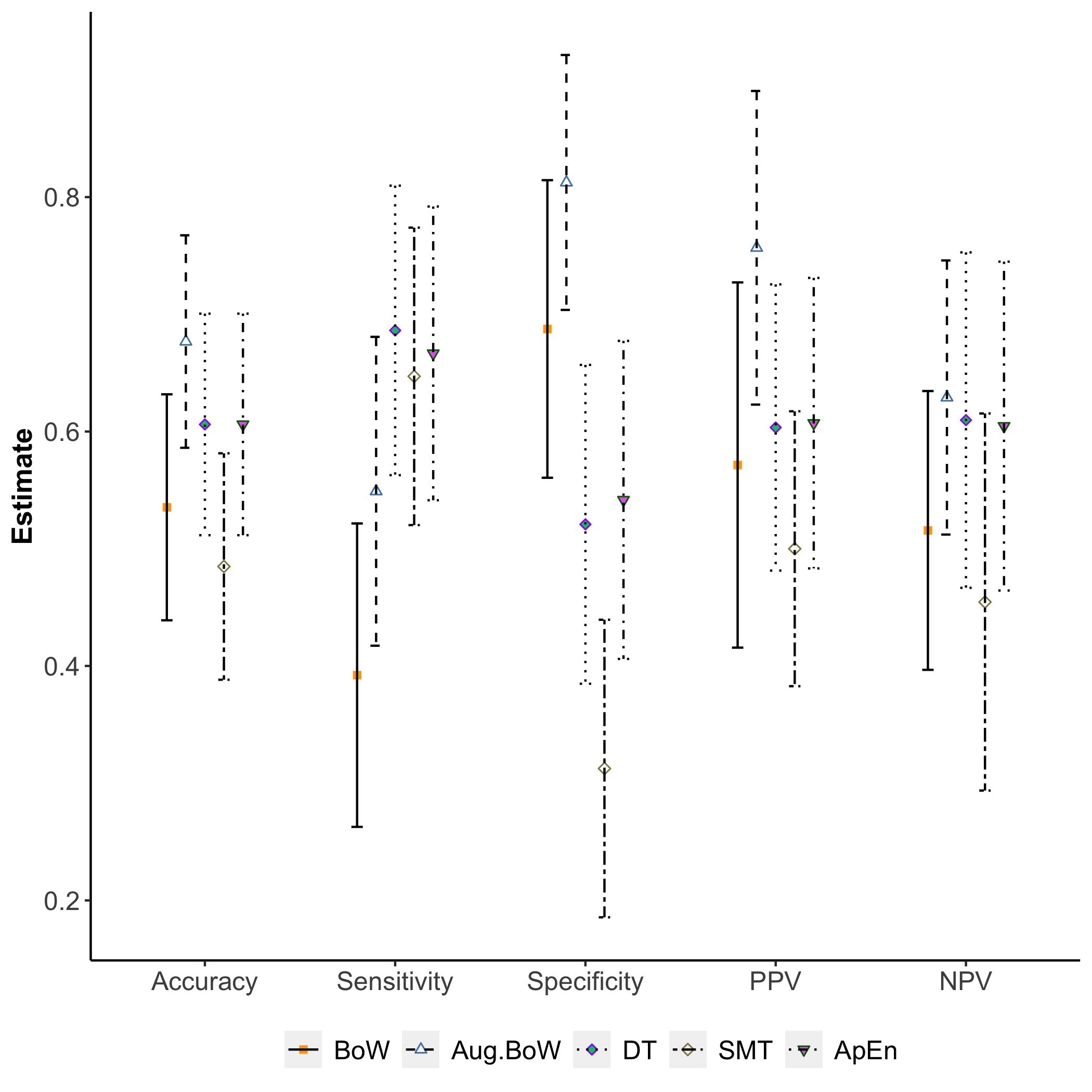}
  \end{center}
  \caption{Predictive performance of interest-point based methods.}
  \label{fig:err_cvlinsvm}
\end{figure}

\begin{table}[]
\begin{minipage}{\textwidth}
\caption{Estimates of accuracy of deep learning methods for scores on individual items in ICO-OSCAR:phaco for capsulorhexis (95\% confidence intervals in parentheses). KP = predicted keypoints analyzed with a temporal convolutional neural network (TCN); ATT = neural network with attention mechanisms. CF = Commencement of flap \& follow through; RF = formation and completion.}
\label{tab:dlacc}       
\begin{tabular}{@{}l|ccc@{}}
\toprule
\textbf{Item} & \textbf{AUC} & \textbf{Micro accuracy} & \textbf{Macro accuracy} \\ \midrule

\multicolumn{4}{l}{\textbf{KP: predicted keypoints analyzed with a TCN}} \\ \midrule

CF & 0.67 (0.59 to 0.72) & 0.65 (0.73 to 0.55) & 0.76 (0.67 to 0.84)  \\ \midrule

RF & 0.72 (0.63 to 0.78) & 0.60 (0.69 to 0.50) & 0.73 (0.64 to 0.81) \\ \midrule

\multicolumn{4}{l}{\textbf{ATT: neural network with attention mechanisms}} \\ \midrule

CF & 0.71 (0.64 to 0.77) & 0.64 (0.72 to 0.54) & 0.76 (0.67 to 0.83)  \\ \midrule

RF & 0.65 (0.58 to 0.70) & 0.48 (0.58 to 0.39) & 0.66 (0.56 to 0.74)  \\ \bottomrule
\end{tabular}
\end{minipage}
\end{table}

\begin{table}[]
\begin{minipage}{\textwidth}
\caption{Estimates of performance measures for scores on individual items in ICO-OSCAR:phaco for capsulorhexis (95\% confidence intervals in parentheses). KP = predicted keypoints analyzed with a temporal convolutional neural network (TCN); ATT = neural network with attention mechanisms. CF = Commencement of flap \& follow through; RF = formation and completion; PPV = positive predictive value; NPV = negative predictive value.}
\label{tab:dlitems}       
\begin{tabular}{@{}l|cccc@{}}
\toprule
\textbf{Item} & \textbf{Sensitivity} & \textbf{Specificity} & \textbf{PPV} & \textbf{NPV}  \\ \midrule

\multicolumn{5}{l}{\textbf{KP: predicted keypoints analyzed with a TCN}} \\ \midrule

CF = 2/3 & 0.00 (0.00 to 0.22) & 0.99 (0.94 to 1.00) & 0.00 (0.00 to 0.95) & 0.86 (0.77 to 0.91)  \\ \midrule

CF = 4 & 0.64 (0.48 to 0.77) & 0.72 (0.59 to 0.81) & 0.60 (0.44 to 0.73) & 0.75 (0.63 to 0.85)  \\ \midrule

CF = 5 & 0.85 (0.72 to 0.92) & 0.68 (0.55 to 0.79) & 0.70 (0.57 to 0.80) & 0.84 (0.70 to 0.92)  \\ \midrule

RF = 2/3 & 0.35 (0.19 to 0.55) & 0.93 (0.86 to 0.97) & 0.62 (0.36 to 0.82) & 0.83 (0.73 to 0.89)  \\ \midrule

RF = 4 & 0.45 (0.30 to 0.62) & 0.80 (0.69 to 0.88) & 0.54 (0.36 to 0.70) & 0.75 (0.63 to 0.83)  \\ \midrule

RF = 5 & 0.84 (0.70 to 0.92) & 0.61 (0.48 to 0.72) & 0.62 (0.49 to 0.73) & 0.83 (0.69 to 0.91)    \\ \midrule

\multicolumn{5}{l}{\textbf{ATT: neural network with attention mechanisms}} \\ \midrule

CF = 2/3 & 0.07 (0.00 to 0.31) & 0.99 (0.94 to 1.00) & 0.50 (0.03 to 0.97) & 0.87 (0.78 to 0.92)  \\ \midrule

CF = 4 & 0.62 (0.46 to 0.75) & 0.73 (0.61 to 0.83) & 0.60 (0.45 to 0.74) & 0.75 (0.62 to 0.84) \\ \midrule

CF = 5 & 0.83 (0.69 to 0.91) & 0.64 (0.51 to 0.76) & 0.67 (0.54 to 0.78) & 0.81 (0.67 to 0.90)  \\ \midrule

RF = 2/3 & 0.17 (0.07 to 0.37) & 0.87 (0.77 to 0.93) & 0.29 (0.12 to 0.55) & 0.78 (0.68 to 0.85)  \\ \midrule

RF = 4 & 0.36 (0.22 to 0.53) & 0.71 (0.59 to 0.81) & 0.39 (0.24 to 0.56) & 0.69 (0.57 to 0.79)  \\ \midrule

RF = 5 & 0.74 (0.60 to 0.85) & 0.61 (0.48 to 0.72) & 0.59 (0.46 to 0.71) & 0.76 (0.61 to 0.86) \\ \bottomrule
\end{tabular}
\end{minipage}
\end{table}

\section{Discussion}\label{sec4}
In this study, we evaluated multiple conventional feature based methods and deep learning methods to assess surgical skill in the operating room using a common data set, experiment setup, and performance measures. Our findings show that deep learning methods perform better than the feature based methods in terms of AUC (Figures \ref{fig:roc_cvlinsvm} and \ref{fig:roc_dl}). Furthermore, a network using attention mechanisms had the most desirable performance measures for skill assessment directly using RGB videos of the surgical field. Even when compared with a network trained using precise manual annotations of instrument tips \citep{kim2019objective}, we observed higher sensitivity (0.843 vs. 0.824) and specificity (0.75 vs. 0.71) with the attention-based network. These findings not only indicate that precise annotations of instrument motion may not be necessary but also that it is useful to analyze the entire context in the surgical field to assess skill instead of analyzing instrument motion alone. It appears that deep learning is the way forward to develop algorithms for video-based assessment of intraoperative surgical skill. Among the conventional feature based methods, Aug.BoW showed higher measures of performance. However, external validity of deep learning methods and Aug. BoW should be determined with data from multiple surgeons at different hospitals.

As the state-of-the art on algorithms for video-based assessment of surgical skill in the operating room, this work shows feasibility and evidence of internal validity. Past studies using videos to assess surgical skill were limited to the simulation setting \citep{zia2018automatedpaper,zia2018video,ahmidi2017dataset,vedula2017objective}. Most studies on algorithms for surgical skill assessment in the operating room used instrument motion \citep{vedula2017objective}. It is clear that instrument motion is informative for surgical skill assessment. In fact, our prior work using surgical video relied upon manual annotation of instrument tips in the images \citep{kim2019objective}. Our findings from analysis of predicted instrument tips (KP) in this study reinforce our prior observation that instrument motion in capsulorhexis can be used to discriminate surgical skill. However, performance of algorithms using predicted keypoints were lower than that obtained from using precise manual annotations (AUC 0.79 vs. 0.86) \citep{kim2019objective}. While it is likely that larger data sets can improve accuracy in predicted keypoints, and subsequently in skill assessment, future research should consider methods to encode additional context in the surgical field that is not limited to instrument motion.

Among the feature based methods, Aug. BoW, a simple method to analyze overall temporal information, performed better than the other methods. This is to be anticipated because Aug. BoW involves encoding of overall temporal information either at a low- or high-level depending on the type of encoding. On the other hand, DT, SMT, and ApEn involve extracting specific types of temporal features. This specificity limits their utility in a complex real-world environment of the operating room as opposed to a controlled simulated environment in which these methods were developed \citep{zia2018automatedpaper,zia2018video}. Furthermore, large variances in estimates of measures of discrimination for these methods, observed in our study, indicate that more data may be necessary to learn a useful classifier with the high-dimensional features. This means that estimates of performance of the feature based methods in our study are likely a consequence of the classifier over-fitting to the features. Despite the possibility that larger data sets may result in more accurate feature based methods, there is little in our findings to suggest that they will generalize to other data sets and be useful for surgical skill assessment in the operating room. Moreover, deep learning methods may lead to more potent discriminative classifiers when larger data sets are available given their end-to-end training.

Performance of algorithms for video-based assessment should be evaluated in the context of the intended application. Video-based assessment of surgical skill in the operating room has multiple applications, each with different stakes or consequences. Surgical skill is associated with patient outcomes, therefore, interventions to improve surgeons' skill can advance quality of care. Skill assessment is key to training surgeons. It supports deliberate practice, and provides summative evaluation at the end of rotations, at the end of each year of training, etc. Some high-stakes applications of surgical skill assessment include certification and re-certification of surgeons for independent practice, determination of operating privileges, and end of career decisions. Not all applications of video-based assessment of surgical skill require the same algorithm performance profile. For example, applications with significant consequences of a false positive, such as certification of surgeons, may require higher specificity (and NPV) than sensitivity (and PPV). In fact, deep learning algorithms in our study, if shown to be externally valid, may not have sufficient specificity for high-stakes assessments, but they may be useful for routine training curricula.

Regardless of the application, to be useful for surgeons to acquire skill, algorithms for surgical skill assessment should be valid and feasible. Consequently, ideal algorithms for video-based surgical skill assessment capture relevant information in video images instead of extraneous patterns in the data. This is typically analyzed by evaluating algorithms in external data sets or different procedures. While generalizability to other surgical procedures is desirable, external validity in other data sets of the same procedure is essential. Such studies should consider not only validity but also feasibility in terms of robustness to heterogeneity in data sources or data collection protocols (e.g., video resolution), and use cases.

Our study has a few limitations, besides a limited amount of data, that provide directions for future research. Our analyses did not account for multiple videos from the same surgeons. For prediction of scores on individual assessment items, imbalance in distribution of groundtruth labels may have adversely affected algorithm performance. We did not analyze sensitivity of algorithm performance to video resolution. Capsulorhexis is performed through microscopic surgical actions, therefore, videos with a high resolution may enable analysis of granular data patterns. In terms of attention mechanisms for skill assessment, our investigation did not include an analysis of the validity and correctness of the learned attention maps. To utilize the learned attention maps for an interpretable assessment of skill and generation of actionable feedback for the surgeon, a qualitative analysis of the usability, feasibility, and correctness of attention maps must be provided. Though our approach uses attention maps to implicitly localize relevant parts of the video without verification, a structured analysis of the acquired attention maps is an interesting direction for future research.

Our work, and other research on algorithms for surgical skill assessment in the operating room, is constrained by non-technical reasons. In particular, there are no known prespecified minimal benchmarks that performance of algorithms for intraoperative surgical skill assessment should achieve. Such benchmarks depend upon the application or use case for the algorithms. While regulatory agencies typically set these benchmarks for medical devices used in patient care, including software as a medical device, \citep{carroll2016software}, it is up to the surgical community to specify them for algorithms for surgical skill assessment in the operating room. These benchmarks are necessary to evaluate the impact of false positive and false negative algorithm predictions in an appropriate context. It is unlikely for any method to always return the correct prediction; thus, understanding incorrect predictions in the context of their application can ensure identifying methods that are useful and avoid discarding methods altogether.

\section{Conclusion}\label{sec5}
Deep learning methods are necessary for video-based assessment of surgical skill in the operating room. While our findings show internal validity of deep learning methods for this purpose in capsulorhexis, testing them in additional data sets is necessary to establish their external validity.

\section*{Declarations}
\begin{itemize}
\item \textbf{Competing Interests:} None.
\item \textbf{Acknowledgments:}
Dr. Austin Reiter, Assistant Research Professor at Johns Hopkins University, mentored this work in its early stages. 
\item \textbf{Funding:} Drs. Vedula, Sikder, and Hager are supported by a grant from the National Institutes of Health, U.S.A.; NIHR01EY033065. The content is solely the responsibility of the authors and does not necessarily represent the official views of the National Institutes of Health.
\end{itemize}

\bibliographystyle{sn-basic}
\bibliography{refs}  






\end{document}